\documentclass[final,5p,twocolumn,times,authoryear]{elsarticle}

\usepackage{graphicx}
\usepackage{booktabs}
\usepackage{amsmath,amssymb,amsfonts}%
\usepackage{amsthm}%
\usepackage{mathrsfs}%
\usepackage{textcomp}%
\usepackage{manyfoot}%
\usepackage{booktabs}%
\usepackage{algorithm}%
\usepackage{algorithmicx}%
\usepackage{algpseudocode}%
\usepackage{listings}%
 \usepackage{makecell}
\usepackage{multirow}
 \usepackage{caption}
\usepackage{subcaption}
\usepackage{enumitem}
\usepackage{textcomp}

\usepackage[breaklinks,colorlinks]{hyperref}

\usepackage[capitalize]{cleveref}
\crefname{section}{Sec.}{Secs.}
\Crefname{section}{Section}{Sections}
\Crefname{table}{Table}{Tables}
\crefname{table}{Tab.}{Tabs.}
\Crefname{figure}{Figure}{Figures}
\crefname{figure}{Fig.}{Figs.}

\begin{document}
\makeatletter
\def\ps@pprintTitle{%
   \let\@oddhead\@empty
   \let\@evenhead\@empty
   \let\@oddfoot\@empty
   \let\@evenfoot\@empty
}
\makeatother
\begin{frontmatter}



\title{GADS: A Super Lightweight Model for Head Pose Estimation} 

\author[label1]{Menan Velayuthan}
\ead{vmenan95@gmail.com}
\cortext[cor1]{Corresponding author}
\author[label1]{Asiri Gawesha\corref{cor1}}
\ead{asiri.l@sliit.lk}
\author[label2]{Purushoth Velayuthan}
\author[label1]{Nuwan Kodagoda}
\author[label1]{Dharshana Kasthurirathna}
\author[label1]{Pradeepa Samarasinghe}
\affiliation[label1]{organization={Faculty of Computing, Sri Lanka Institute of Information Technology},
            addressline={New Kandy Road},
            city={Malabe},
            postcode={10115},
            state={Western},
            country={Sri Lanka}}

\affiliation[label2]{organization={Dept. of Computer Science \& Engineering, University of Moratuwa},
            addressline={Bandaranayake Mawatha,},
            city={Moratuwa},
            postcode={10400},
            state={Western},
            country={Sri Lanka}}

\begin{abstract}
In human-computer interaction, head pose estimation profoundly influences application functionality. Although utilizing facial landmarks is valuable for this purpose, existing landmark-based methods prioritize precision over simplicity and model size, limiting their deployment on edge devices and in compute-poor environments. To bridge this gap, we propose \textbf{Grouped Attention Deep Sets (GADS)}, a novel architecture based on the Deep Set framework. By grouping landmarks into regions and employing small Deep Set layers, we reduce computational complexity. Our multihead attention mechanism extracts and combines inter-group information, resulting in a model that is $7.5\times$ smaller and executes $25\times$ faster than the current lightest state-of-the-art model. Notably, our method achieves an impressive reduction, being $4321\times$ smaller than the best-performing model. We introduce vanilla GADS and Hybrid-GADS (landmarks + RGB) and evaluate our models on three benchmark datasets---AFLW2000, BIWI, and 300W-LP. We envision our architecture as a robust baseline for resource-constrained head pose estimation methods.
\end{abstract}





\begin{keyword}
Head Pose Estimation \sep Efficient ML \sep Geometric Deeplearning \sep Multi-head attention



\end{keyword}

\end{frontmatter}



\section{Introduction}\label{introduction}

Head Pose Estimation (HPE) is an essential tool used in numerous apps and cutting-edge technologies. With the dawn of data-driven solutions, interest in human behavior analysis and facial emotion detection~\citep{kepler, FAN} has taken center stage in research. Human behavior analysis is a multidisciplinary field that involves the study of human actions, movements, and interactions in different contexts~\citep{humanPose,ehpe,mdfnet}. The main purpose of such a study is to understand and interpret human behavior patterns, emotions, and intentions and their relationships to the environment and other individuals. The head and facial regions are the primary sources of information about a person’s emotional state. Therefore, they are crucial factors in understanding human behavior. They are responsible for expressing a wide range of emotions, conveying non-verbal cues, and indicating the direction of attention. Head pose is a good indicator of human attention and is a vital and common pre-processing step in various domains, including human-computer interaction, driver safety, sports analysis, surveillance, and security. HPE has found some direct applications in virtual and augmented reality, where the system creates a more immersive experience by accurately tracking the user's head movements and adjusting the display accordingly.

HPE approaches fall into two main categories~\citep{evagcn}: 1) Landmark-based methods, which rely on facial landmarks to estimate head pose, and 2) Landmark-free methods, which use the images themselves for head pose estimation. Recently, landmark-based methods have not been among the SOTA models, as estimating landmarks is computationally demanding and results in larger, more resource-intensive models. However, advances have led to mobile-friendly landmark detectors that operate in near real-time~\citep{mediapipe}. Additionally, applications like virtual reality~\citep{kepler}, expression transfer~\citep{wang2018every}, and face alignment~\citep{FAN} already incorporate landmark estimation as a precursor for downstream tasks, making these landmarks readily available as metadata for HPE. Our primary goal is to develop the lightest and fastest model that can compete with robust SOTA models for head pose estimation. Achieving this goal has allowed us to shift the workflow's computational bottleneck to the landmark estimator, as our model introduces minimal latency to the pipeline.

We introduce \textbf{``Grouped Attention Deep Sets (GADS)"}, a mobile-friendly model designed to match the competitive estimation capabilities of existing SOTA methods. We assert that the proposed architecture's utility is not confined solely to the domain of head pose estimation; it's applicability extends to any problem involving Landmark-based analysis.

In our previous work, DS-HPE~\citep{dshpe}, we employed the Deep Set architecture~\citep{manzil2017deepset}, allowing us to conceptualize landmarks as a set of points. Deep Set is a model designed for conducting deep learning on sets, with it's distinctive permutation-invariant property enabling the representation of landmarks as a ``set".

Although DS-HPE demonstrated comparable performance against SOTA models, it utilized the complete set of landmarks, potentially neglecting information unique to specific facial regions (e.g., eyes, cheeks, chin) by focusing on the entire set simultaneously. Therefore, in this study, we utilize Deep Set as a layer and organize groups for each distinct region of the face (right eye, left eye, left cheek, right cheek, and chin). Additionally, we incorporate ``multihead-attention"~\citep{vaswani2017attention} to capture intergroup relationships.

Our results indicate that our model performs competitively with SOTA models while maintaining a size that is $7.5\times$ smaller than the smallest model documented in the literature. To the best of our knowledge, this is the first instance of Deep Set being utilized in conjunction with multihead attention for a Head Pose Estimation (HPE) task. Furthermore, we introduce a hybrid model (GADS-Hybrid) that takes both images and landmarks as input to predict yaw, pitch, and roll.

Head pose is a vector in 3D space with dimensions of yaw, pitch, and roll. There have been numerous work that utilize 3D information using depth images~\citep{xiao2020,borghi2018face,borghi2017poseidon}, but they come with the overhead of using depth cameras and other complex technologies in practice, which are expensive to use. It is ideal to use 3D information instead of 2D landmarks, without the additional costs and computational overheads. Recent advances in computer vision have paved the way for a variety of methods that can estimate 3D landmarks~\citep{mediapipe,FAN} from 2D images.

In this study, we propose a novel head pose estimation model named GADS, which is based on the Deep Set architecture with a self-attention mechanism. We show that despite it's small size, our proposed model achieves competitive accuracy compared to state-of-the-art models. GADS is also the lightest model implemented for head pose estimation, with potential for extension to other problems that utilize landmark coordinates.

Additionally, we introduce a hybrid model architecture that combines GADS with a convolutional neural network (CNN) that employs both landmark coordinates and RGB images of the face.

We evaluate the performance of both GADS vanilla and hybrid architectures using three benchmark datasets, measuring the mean absolute error of the three angles: yaw, pitch, and roll. We compare the predicted values against ground truth and assess the computational time of our proposed architecture against SOTA methods. We also conduct an extensive ablation study to investigate the impact of hyper-parameters on the model's performance, and select the best parameters to provide the highest accuracy with the least number of parameters, keeping the model size as small as possible while still performing competitively with SOTA methods.

This paper is organized as follows: In ~\Cref{related_work}, we provide an overview of various approaches in the literature used for head pose estimation, including the DeepSet model architecture, self-attention, and multi-head attention mechanisms. We introduce our proposed model architectures, GADS vanilla and GADS hybrid, in ~\Cref{methdolology}. There, we provide a detailed explanation of each architecture and include visualizations of their respective structures. In ~\Cref{experiments}, we discuss the datasets used for our experiments, the evaluation protocols, and the implementation details of our model architectures. In ~\Cref{results}, we present the results of our experiments, comparing the performance of our models with SOTA methods using two different protocols. We include visualizations of predicted angles compared to ground-truth values, as well as results from an ablation study and a comparison of execution times between GADS and SOTA methods. Finally, we conclude with a summary of our study.

\section{Related Work}\label{related_work}

\subsection{Head Pose Estimation}

As detailed in ~\Cref{introduction}, head pose estimation can be broadly categorized into Landmark-based and Landmark-free methods. Landmark-based approaches~\citep{evagcn, Dlib} utilize facial landmarks to estimate the head orientation based on their relative positions. Conversely, landmark-free approaches~\citep{kepler, lathuiliere2017deep, albiero2021img2pose, ranjan2018light} focus on directly estimating head pose from the input image without relying on identified facial landmarks.

While Landmark-based approaches have historically achieved SOTA performances, landmark-free methods have gained popularity due to their ability to accurately estimate head pose in challenging conditions and generalize effectively to new datasets. This shift in preference may be attributed to the overhead introduced by pre-processing methods involved in computing and estimating landmarks in Landmark-based approaches.
However, applications such as facial recognition, augmented reality, gaming~\citep{edmf}, and human-computer interactions~\citep{Anisotropic, ngdnet, arhpe} that rely on landmarks for downstream tasks also necessitate head pose estimation for their algorithms. In these cases, where landmarks are readily available, Landmark-based methods can be effectively employed for head pose estimation. Moreover, recent advancements in landmark extraction techniques have resulted in near real-time estimation capabilities, as demonstrated by~\citep{mediapipe}.

Remarkably, the current SOTA lightweight model in the literature is LwPosr~\citep{lwposr}. The author employs a combination of depthwise separable convolutional (DSC) and transformer encoder layers to achieve leading performance while minimizing the model size. The compact size is credited to the efficiency of DSC layers, which consume less memory than standard CNN layers. As far as our knowledge extends, LwPosr stands out as the lightest SOTA model in the literature for Head Pose Estimation.

\subsection{Deep Set and Deep Sets for HPE}
Deep Sets~\citep{manzil2017deepset} represents a specialized architecture designed for deep learning on ``Sets." The key characteristic of this architecture is it's utilization of permutation-invariant functions to capture the inherent set property. Other notable architectures leveraging the set property include PointNet~\citep{qi2017pointnet} and PointNet++~\citep{qi2017pointnet++}, primarily applied in the domain of point clouds~\citep{guo2020deep}.

PointNet and PointNet++ operate on depth images, necessitating the use of RGB-D cameras. However, the limited availability of RGB-D cameras for mobile-based applications poses a challenge for integrating these Head Pose Estimation (HPE) architectures into such systems. In our prior work~\citep{dshpe}, we introduced Deep Set for Head Pose Estimation (DS-HPE). Despite it's simplicity, our model demonstrated competitive performance against SOTA architectures and outperformed them in terms of computational efficiency. However, DS-HPE was not specifically optimized for compute poor environments. This paper aims to address the gaps identified in DS-HPE, with a focus on optimizing the model for compute poor environments.

\subsection{Self Attention and Multi-head Attention}

The attention mechanism~\citep{bahdanau2014neural} was originally introduced for sequence-to-sequence (Seq2Seq) problems as a tool to establish alignment between two sequences. This mechanism assigns weights or importance scores to each element of the source sequence when predicting corresponding elements in the target sequence. Self-attention~\citep{shaw2018self} occurs when a sequence focuses on itself to discern key relationships within the sequence. In this paper, the self-attention mechanism is employed to learn relationships between intermediate vectors.

Multi-Head attention~\citep{vaswani2017attention}, within the context of self-attention, involves stacking multiple self-attention layers in parallel as attention heads. This configuration enables the model to learn different aspects of relationships between elements in the sequence. While the attention mechanism was initially introduced for Neural Machine Translation~\citep{luong2015effective}, it has found applications in various domains such as vision~\citep{dosovitskiy2020image}, speech~\citep{chorowski2015attention}, and others.

In our work, we adopt a self-attention-based multi-head attention approach, similar to that used in~\citep{vaswani2017attention}.

\section{Proposed Method}
\label{methdolology}
\begin{figure}[h!]
    \begin{subfigure}[]{\columnwidth}
      \flushleft
    \includegraphics[width=270pt]{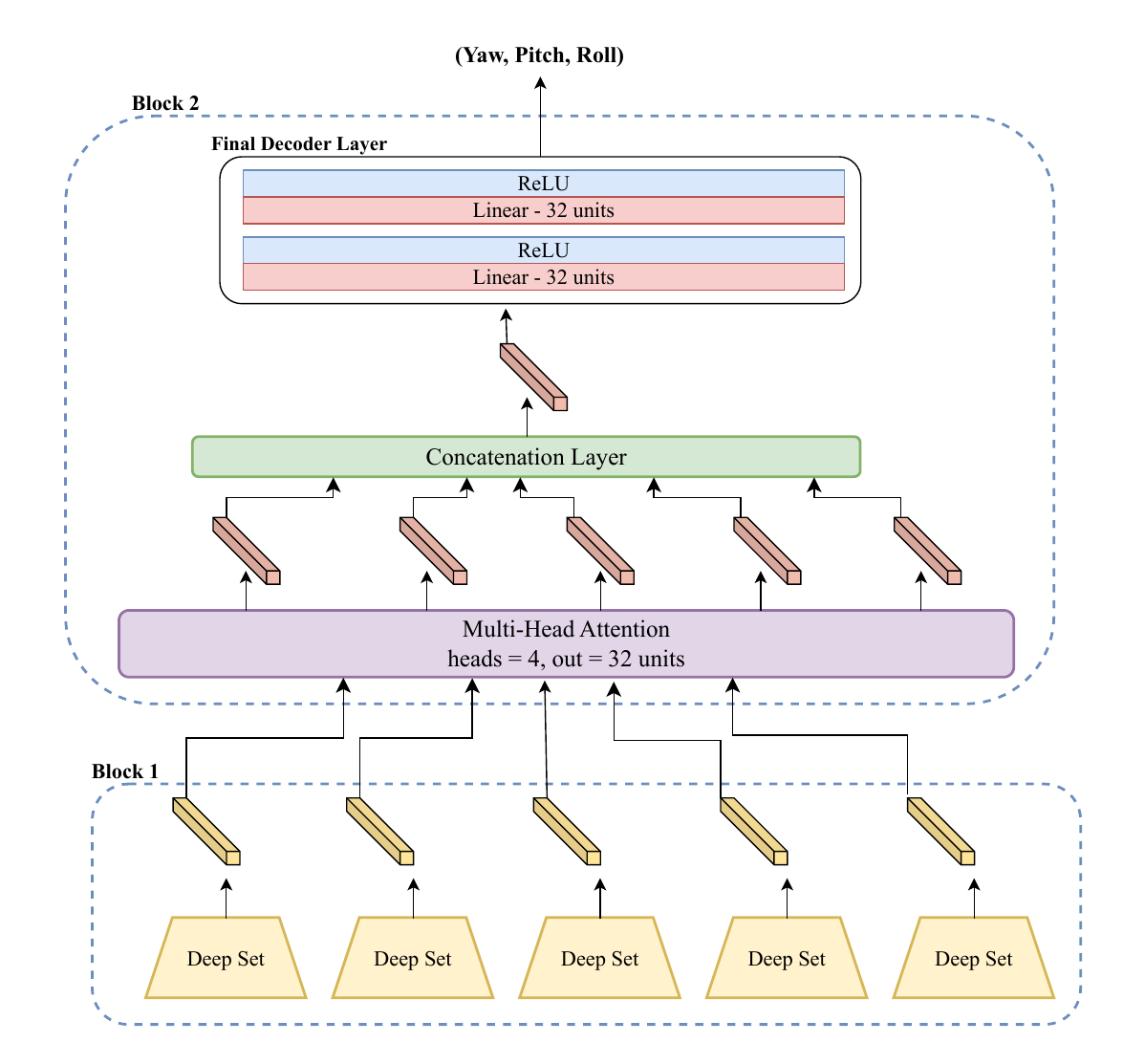}
    \caption{Overview of the vanilla GADS architecture: The five landmark groups enter the Deep Set layer, followed by a Multi-head attention layer with four attention heads, each producing a 32-dimensional vector. The outputs are concatenated and processed through the final Decoder layer, composed of two pairs of Linear + ReLU units, to generate the final output.}
    \label{fig:gads_arch}
     \end{subfigure}
      \begin{subfigure}[]{\columnwidth}
      \flushleft
    \includegraphics[width=270pt]{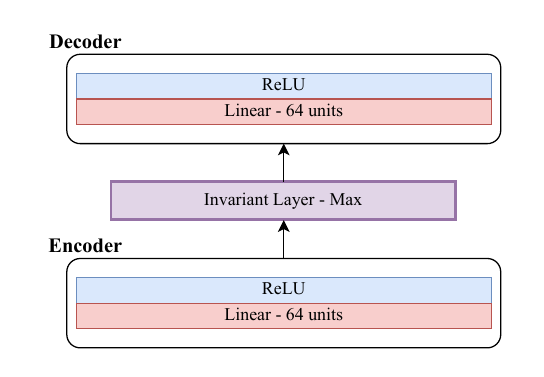}
    \caption{Overview of a single Deep Set layer, which consists of Encoder Layer, Permutation Invariant Operator and Decoder Layer.}
    \label{fig:deepset_arch}
     \end{subfigure}
 \caption{Architectures of the GADS and the Deep Set Layer}

 \label{fig:model_archs}
\end{figure}

This paper follows the same pre-processing steps as our previous work~\citep{dshpe} to obtain the landmarks for the model. Given that the pre-processing steps are shared between the GADS and GADS-Hybrid models, we address them commonly as follows:

As a first step in the data pre-processing, the 3-channel RGB image $I$ is fed into the face detector block. This face detector block is designed to identify potential regions in the input image that contain human faces and generate a set of regions of interest (ROI) corresponding to the detected faces(~\Cref{eq:1}).

\begin{equation}\label{eq:1}
    \ RL = \Gamma(\Lambda(I))
\end{equation} 
        where:
        \begin{description}[noitemsep]
        \item$I$    = raw image,
        \item$\Lambda$  = face detector block,
        \item$\Gamma$   = landmark detector block,
        \item$RL$   = \{$rl_i\ ; i= 1...N $\} where $rl_i$$\in$ $R^3$ is a single raw 3D face landmark coordinate $(x_i, y_i, z_i)$ and $N$ is the total number of landmarks generated for the ROI.
        \end{description}
Next, landmarks are normalized to improve performance of the model. A normalized landmark $l_i$ can be represented by~(\Cref{eq:2}),
\begin{equation}\label{eq:2}
\ l_i = \frac{rl_i - rl_0}{rl_{max} - rl_{min}}
\end{equation}
where:
        \begin{description}[noitemsep]
        \item $rl_i$: set of raw landmarks defined in~(\Cref{eq:1}),
        \item $rl_0$: reference landmark, the new coordinate values will be calculated relative to this point. For the purpose of this research, landmark on the nose is selected as the reference landmark~\citep{evagcn, wedasingha2022skeleton}, 
        \item $rl_{max}, rl_{min}$: maximum and minimum values of the set of landmarks $RL$.
        \end{description}
\hfill\break
As the final pre-processing step, landmarks are grouped according to their identified regions as shown in \Cref{fig:land_sets} (eg: left eye, right eye, chin, left cheek, right cheek).

\begin{equation}\label{eq:3}
\ GP = \{G_i\ ; i= 1...M \}
\end{equation}
             where:
                \begin{description}[noitemsep]
        \item $GP$: the landmarks grouped together by regions they belong to (eg: right eye region, left cheek region etc..),
        \item $G_i$: individual set of landmark regions,
        \item $M$: the total number of groups. With this architecture, we can expand to any number of groups (in this paper we keep M=5). 
        \end{description}
\hfill\break
A summary of the above mentioned steps are to initially, detect the face and crop the ROI. Second, to detect the landmarks for the ROI. Next, is to normalize the landmarks from ROI. Finally, group the landmarks.

In the following two subsections~\Cref{gads_vanilla_model} and~\Cref{gads_hybrid_section} we will discuss our proposed two model architectures: GADS vanilla model and the GADS-Hybrid model.

\subsection{GADS Vanilla Model}
\label{gads_vanilla_model}

The GADS architecture can be divided into two main blocks:
\begin{itemize}[noitemsep]
    \item Block 1: contains parallel Deep Set~\citep{manzil2017deepset} layers. Here each $GP$(~\Cref{eq:3}) has a respective deep set layer dedicated for it. The number of Deep Set Layers depends on the number of groups present.
    \item Block 2: contains a multi-head attention layers~\citep{vaswani2017attention}, a concatenation layer and a final decoder layer.
\end{itemize}
The architecture and the blocks are shown in ~\Cref{fig:gads_arch}. 

\subsubsection{Block 1: Parallel Deep Set Layers}
Each group, $G_i$~(\Cref{eq:3}) has a dedicated layer for itself. Hence the name ``parallel'' deepset layer. The number of parallel layers is controlled by the number of groups we choose, in this work the number of parallel layers is five. A simple overview diagram of the Deep Set layer is shown in ~\Cref{fig:deepset_arch}. A deep set layer can be defined as:

\begin{equation}
\ H_i = f(G_i)
\end{equation}
             where:
                \begin{description}[noitemsep]
            \item$f$: encoder,
            \item $G_i$: $i^{th}$ group of $GP$(~\Cref{eq:3}),
            
            \item$H_i$: \{$h_j\ ; i= 1...S $\} where $h_j \in R^d$ is the latent representation in $d$-dimensional embedding space. $S$ is the total number of landmarks per group (in our work $S$ is either 5 or 6).
        \end{description}
Once the embedded set $H_i$ is obtained, we then perform an invariant combination of the embedding present in the embedding set.
        \begin{equation}
         \ z_i = {{\oplus_{j=1}^{S}} h_j} 
        \end{equation}
             where:
                \begin{description}[noitemsep]
             \item{$\oplus{}$}: placeholder for invariant function eg: summation, average, minimum, maximum,
            
            \item{$z_i$}: belongs to the same embedding space i.e.\ {$z_i \in R^d$}.
        \end{description}
\hfill\break
We then pass $z$ through a decoder to arrive to our final output,
\begin{equation}\label{eq:6}
 \hat{z_i} = g(z_i)
\end{equation}
         where:
                \begin{description}[noitemsep]
                \item$g$: decoder,
                \item$\hat{z_i}$: the latent representation of a group $G_i$.
        \end{description}
\hfill\break
It is a key to note that there is a $\hat{z_i}$ for each deep set layer in the model. In our case there are be five $\hat{z_i}$ for five parallel deep set layers. 

\begin{equation}
  \hat{Z} =  \{\hat{z_i}\ ; i= 1...M \}
\end{equation}
             where:
                \begin{description}[noitemsep]
            \item $\hat{Z}$: collection all the $\hat{z_i}$ from each deep set layer.
            \item $\hat{z_i}$: the latent representation of a group $G_i$~(\Cref{eq:6}).
            \item $M$: the total number of deep set layers. In our case it's 5.
        \end{description}
\hfill\break
The output of Block 1 ($\hat{Z}$) is taken as input by Block 2. Next we will address Block 2 in detail.

\subsubsection{Block 2}
This section describes the block containing the Multi-Head attention layer, concatenation layer and the final decoder layer. For Multi-Head attention our work follows the methodology introduced by~\citep{vaswani2017attention}.

Scaled Dot-Product attention is defined to be,
\begin{equation}
  Attention(Q, K, V) =  softmax(\frac{QK^T}{\sqrt{d_k}})V
\end{equation}
Here, $Q,\ K,\ V$ are Query, Key and Value respectively and $\frac{1}{\sqrt{d_k}}$ is the scaling factor (usually the number of dimensions of Key). Multi-head attention is defined by~\citep{vaswani2017attention} as,

\begin{equation}
  MultiHead(Q, K, V) = Concat(head_1,...,head_h)W^O
\end{equation}
\begin{equation}
      head_i = Attention(QW_i^Q,KW_i^K,VW_i^V)
\end{equation}
     where:
                \begin{description}[noitemsep]
        \item $W_i^Q,W_i^K,W_i^V,W^O$: are projection parameter matrices. Follow~\citep{vaswani2017attention} further details.
        \item $head_i$: single attention head. In our case we use 4 attention heads.
        \end{description}
\hfill\break
It is important to note that since our work uses only self attention~\citep{vaswani2017attention}, we initialize $Q,K,V$ to be $\hat{Z}$. 
\newline
Finally the output of the Multi-head attention(which is transformed $\hat{Z}$) is concatenated to form a single vector. This vector is later transformed by the decoder layer to produce a (yaw, pitch, roll).
\begin{equation}
  E =  MultiHead(\hat{Z}, \hat{Z}, \hat{Z})
\end{equation}
where:
                \begin{description}[noitemsep]
        \item $E$: \{$e_i\ ; i= 1...M $\}, there is $e_i$ for every $\hat{z_i}$(~\Cref{eq:6}).
        \end{description}
\begin{equation}
  \hat{e} =  \|_{i=1}^M e_i
\end{equation}
where:
                \begin{description}[noitemsep]
        \item $\hat{e}$: all of $e_i$ concatenated together,
        \item $\|$: concatenation operator. 
        \end{description}

\begin{equation}
  \hat{y} =  D_{GADS}(\hat{e})
\end{equation}
where:
                \begin{description}[noitemsep]
        \item $\hat{y}$: predicted yaw, pitch and roll,
        \item $D_{GADS}$: final decoder layer.
        \end{description}
\hfill\break    
To make GADS robust, we introduce a Hybrid model, which utilizes a convolution neural network block as detailed in ~\Cref{gads_hybrid_section}. 
\begin{figure}[h!]
    \centering
    \includegraphics[width=270pt]{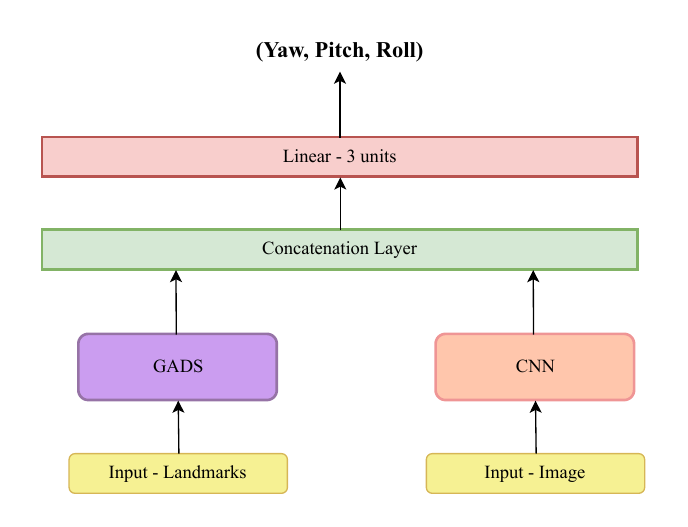}
    \caption{GADS Hybrid Architecture: Landmarks are processed by the vanilla GADS model, while the input image undergoes a CNN block. The outputs are concatenated and fed through a final 3-unit Linear layer to obtain the final output.}

    \label{fig:gads_hybrid_arch}
    
\end{figure}

\subsection{GADS Hybrid Model}
\label{gads_hybrid_section}

\begin{figure}[h!]
    
    \includegraphics[width=270pt]{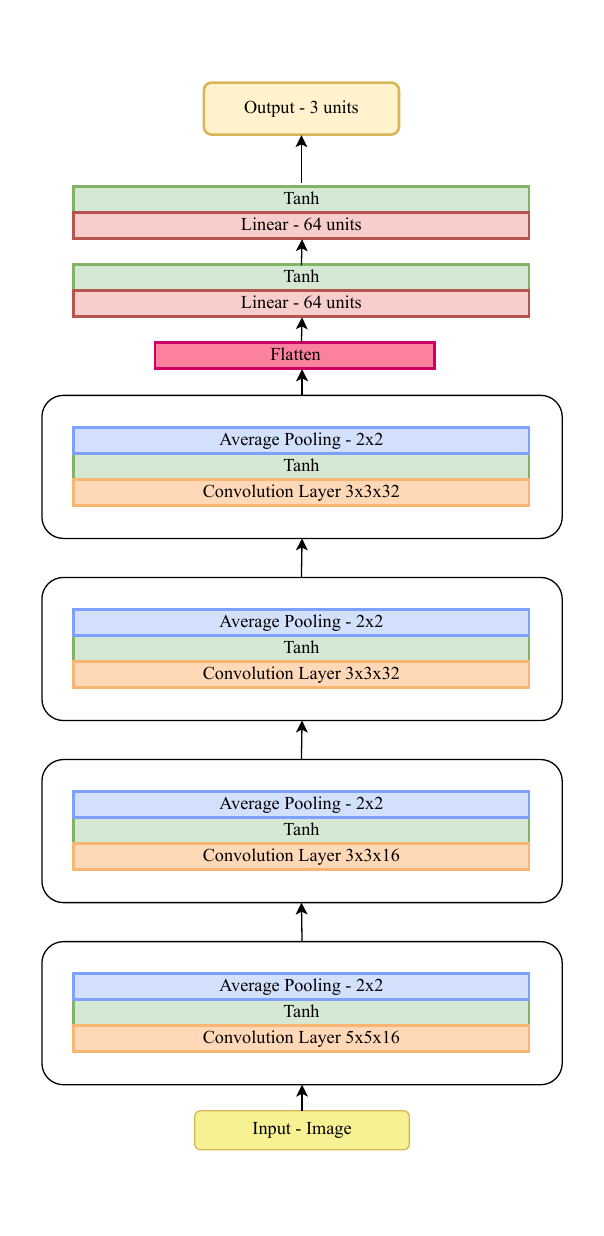}
    \caption{The GADS Hybrid CNN consists of three convolution blocks with $5\times5\times16$ filters, followed by Tanh activation and $2\times2$ Average pooling. The final output is obtained by flattening the last convolution block's output and passing it through two pairs of Linear + Tanh layers.}
    \label{fig:cnn_arch}
\end{figure}

In addition to our vanilla GADS model, we obtain output from a Lenet~\citep{lecun1998gradient} inspired Convolution Neural Network (CNN)~\citep{gu2018recent} by passing the raw image $I$ through the CNN model. The architecture of our hybrid model is illustrated in the~\Cref{fig:gads_hybrid_arch}.
\begin{equation}
      j_1 = GADS(GP) 
\end{equation}
\begin{equation}
      j_2 = CNN(I) 
\end{equation}
Here, $GP$(~\Cref{eq:3}) and $I$(~\Cref{eq:1}) take the same meaning as they are defined previously. Next, $j_1$ and $j_2$ are concatenated to a single vector $\hat{j}$ (~\Cref{eq:16}),
\begin{equation}\label{eq:16}
      \hat{j} = j_1\  \| \ j_2
\end{equation}
where ``$\|$'' stands for concatenation operation.
\begin{equation}
  \hat{y} =  D_{Hyb}(\hat{j})
\end{equation}
     where:
                \begin{description}[noitemsep]
        \item $\hat{y}$: predicted yaw, pitch and roll,
        \item $D_{Hyb}$: final decoder layer .
        \end{description}
\hfill\break
Finally, the concatenated vector $j$ is transformed by a decoder layer to produce $(yaw, pitch, roll)$ values.
\begin{figure*}[h!]
    \centering
    \captionsetup{justification=centering}
    \includegraphics[width=\textwidth]{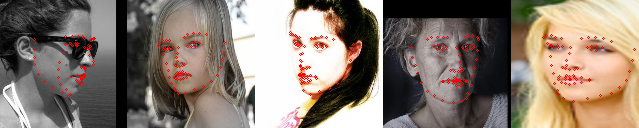}
    \caption{Examples of 68 landmarks extracted using the FAN landmark detector on the AFLW2000 dataset.}
    \label{fig:all_lands}
\end{figure*}

\begin{figure*}[h!]
    \centering

    \includegraphics[width=\textwidth]{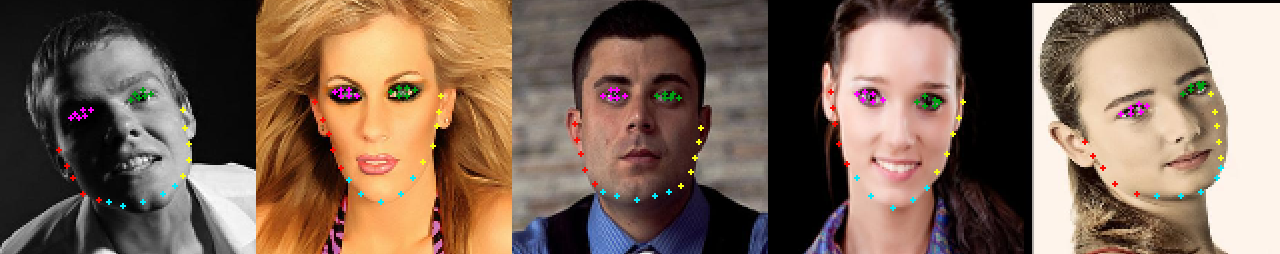}
    \caption{Illustration of 27 selected landmarks grouped into five sections, each represented by a distinct color. Landmarks within the same color belong to the same group: Left eye (6 points) - Purple, Right eye (6 points) - Green, Left cheek (5 points) - Red, Right cheek (5 points) - Yellow, Chin (5 points) - Cyan.}
    \label{fig:land_sets}
\end{figure*}

\subsection{Data-sets}
To train and evaluate our proposed method against SOTA methods, we employ three challenging datasets: 300W-LP~\citep{300WLP}, AFLW2000~\citep{300WLP}, and BIWI~\citep{BIWI}, each annotated with yaw, pitch, and roll angles in 3D.

\textbf{BIWI dataset} comprises 15,768 annotated yaw, pitch, and roll head poses. \textbf{300W-LP} is a compilation of 61,225 samples extracted from four subsets: \textbf{AFW}~\citep{AFW}, \textbf{LFPW}~\citep{LPW}, \textbf{HELEN}~\citep{HELEN}, and \textbf{IBUG}~\citep{IBUG}. Augmentation through flipping resulted in a total of 122,450 synthesized records. \textbf{AFLW2000} includes the initial 2000 records from the AFLW~\citep{AFW-ori} dataset. These three datasets serve as the basis for evaluating our proposed methods and comparing them with SOTA approaches. Mean Absolute Error (MAE) is utilized as the evaluation metric.

Additionally, human faces are detected and extracted using the MTCNN~\citep{mtcnn} face detector, consistent with previous SOTA methods such as EVAGCN~\citep{evagcn}, FSANET~\citep{fsa}, and LwPosr~\citep{lwposr}. No image loss occurred when processing images through the face detector for any of the three datasets in this study. We resize the cropped face areas of the images to $64 \times 64$, maintaining consistency across all experiments.

Since both of our approaches are Landmark-based, we employ the FAN landmark detector~\citep{FAN} to extract landmarks across all frames of the three datasets, providing 68 3D landmarks for each frame. This ensures consistency in our evaluation criteria with other SOTA methods. Frames lost due to occlusions, warped images, and extremely high angles during processing through the landmark detector are summarized in ~\Cref{tab:frame_loss}. A selection of samples from each dataset is visualized to observe the distribution of the 68 landmarks for different head pose angles(~\Cref{fig:all_lands}).

\subsection{Experimental protocols}
We evaluate our proposed architecture against previous methods in the literature, maintaining consistency by following the same evaluation protocols as FSANet~\citep{fsa}, Eva-GCN~\citep{evagcn}, HopeNet~\citep{hopenet}, and LwPosr\citep{lwposr}. We have two protocols on-par with he previous studies. 

\textbf{Protocol P1:} Aligning with previous studies, we conduct a comparative analysis by evaluating our two proposed methods, namely Vanilla GADS and the GADS-Hybrid model, against existing approaches trained on the 300W-LP dataset~\citep{300WLP}. The evaluation is performed on the BIWI dataset~\citep{BIWI} and the AFLW2000 dataset~\citep{300WLP}, employing consistent methodologies to ensure a rigorous and fair comparison in the assessment of performance.

\textbf{Protocol P2:} As the second protocol, we trained evaluated our proposed methods and existing methods on BIWI data-set\citep{BIWI} with a 70:30 train-test split, similar to the previous studies~\citep{fsa,evagcn,hopenet,lwposr}. 
 improve this

\subsection{Evaluation Metrics}
\begin{table}[h!]
\centering
\caption{Number of frames and the percentage of the total frames that were lost when images were passed through the FAN landmark detector.}
\label{tab:frame_loss}
\begin{tabular}{l|l|l|l}
\hline
\textbf{Dataset} & \textbf{Before} & \textbf{After} & \textbf{Loss \%} \\ \hline
BIWI             & 15678           & 14954          & 4.62             \\ 
300WLP           & 122450          & 105484         & 13.86            \\ 
AFLW             & 2000            & 1869           & 6.55             \\ 
\end{tabular}
\end{table}
To assess the performance of our model and benchmark it against SOTA methods, we employ the Mean Absolute Error (MAE). Given that we are measuring performance for three angles—yaw, pitch, and roll—we calculate individual absolute errors for each angle. The Mean Absolute Error is then determined by averaging the errors of the three angles. The Mean Absolute Error (MAE) for yaw ($y$), pitch ($p$), and roll ($r$) angle errors for a single observation is defined as follows:

\begin{equation}
    \text{MAE} = \frac{1}{3} \left(|y - \hat{y}| + |p - \hat{p}| + |r - \hat{r}|\right)
\end{equation}

where:
\begin{align*}
    & y, p, r \quad \text{are the actual yaw, pitch, and roll angles}, \\
    & \hat{y}, \hat{p}, \hat{r} \quad \text{are the predicted yaw, pitch, and roll angles}.
\end{align*}

Furthermore, we evaluated execution times in milliseconds to compare the performance of the SOTA methods on both CPU and GPU. Additionally, we compared the number of parameters in the models to assess the size of each model.

\subsection{Implementation Details}
\begin{figure*}[tb]
    \centering
    \includegraphics[width=450pt]{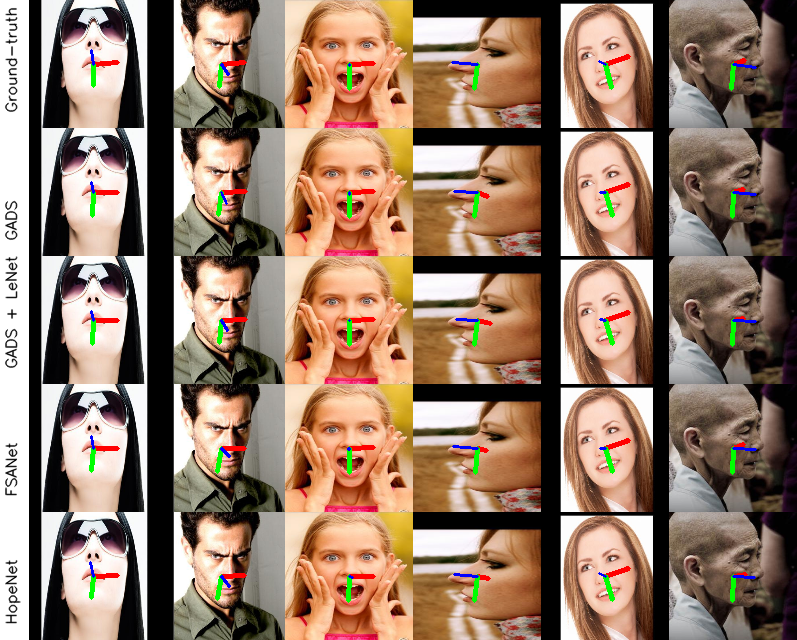}
    \caption{Illustration of yaw, pitch, and roll angles for 5 instances of the AFLW200 dataset. The first row illustrates the correct or ground-truth angles, the second and third rows show the predicted angles of our proposed methods, GADS and GADS-hybrid model, while the fourth and fifth rows display the predicted angles of FSANET and HopeNet. }
    \label{fig:angles_aflw}
\end{figure*}

\section{Experiments}\label{experiments}
\begin{table*}[tb]
\centering
\caption{Mean absolute errors (MAE) and number of parameters for Protocol 1:Trained on the 300W-LP data-set and tested on the BIWI and AFLW2000 data-sets (values of SOTA methods are obtained from the study LwPosr~\citep{lwposr}). Methods are sorted in the descending order based on the number of parameters.  }
\label{tab:Protocol:1}
\begin{tabular}{l|l|cccc|cccc}
\hline
\textbf{} & \textbf{} & \multicolumn{4}{c}{\textbf{BIWI}} & \multicolumn{4}{|c}{\textbf{AFLW2000}} \\
\hline
\textbf{Method} & \textbf{Param $10^6$} & \textbf{Yaw} & \textbf{Pitch} & \textbf{Roll} & \textbf{MAE} & \textbf{Yaw} & \textbf{Pitch} & \textbf{Roll} & \textbf{MAE} \\
\hline
KEPLER & - & 8.80 & 17.30 & 16.20 & 13.90 & - & - & - & - \\
3DDFA & - & 36.20 & 12.30 & 8.78 & 19.10 & 5.40 & 8.53 & 8.25 & 7.39 \\
TokenHPE &86.42 & 4.51 & \textbf{3.95} & \textbf{2.71} & \textbf{3.72} & 5.54 & \textbf{3.36} & 4.08 & 4.66 \\
6DRepNet &39.32 & 3.63 & 4.91 & 3.37 & 3.97 & \textbf{3.24} & 4.48 & \textbf{2.68} & \textbf{3.47} \\
FAN(12 points) & 36.60 & 6.36 & 12.30 & 8.71 & 9.12 & 8.53 & 7.48 & 7.63 & 7.88 \\
TriNet & 26.00 & 4.11 & 4.76 & 3.05 & 3.97 & 4.04 & 5.78 & 4.20 & 4.67 \\
Shao & 24.60 & 4.59 & 7.25 & 6.15 & 6.00 & 5.07 & 6.37 & 4.99 & 5.48 \\
HopeNet(a=1) & 23.90 & 4.81 & 6.61 & 3.27 & 4.90 & 6.92 & 6.64 & 5.67 & 6.41 \\
HopeNet(a=2) & 23.90 & 5.12 & 6.98 & 3.39 & 5.18 & 6.47 & 6.56 & 5.44 & 6.16 \\
Dlib (68 points) & 6.24 & 16.80 & 13.80 & 6.19 & 12.20 & 23.10 & 13.60 & 10.50 & 15.80 \\
WHENet & 4.40 & 3.99 & 4.39 & 3.06 & 3.81 & 5.11 & 6.24 & 4.92 & 5.42 \\
EVA-GCN & 3.30 & 4.46 & 5.34 & 4.11 & 4.64 & 4.01 & 4.78 & 2.98 & 3.92 \\
FSA-Caps-Fusion & 1.20 & 4.27 & 4.96 & 2.76 & 4.00 & 4.50 & 6.08 & 4.64 & 5.07 \\
SSR-NET-MD & 0.20 & 4.49 & 6.31 & 3.61 & 4.65 & 5.14 & 7.09 & 5.89 & 6.01 \\
LwPosr & 0.15 & 4.11 & 4.87 & 3.19 & 4.05 & 4.80 & 6.38 & 4.88 & 5.35 \\
LwPosr-a & 0.15 & 4.41 & 5.11 & 3.24 & 4.25 & 4.44 & 6.06 & 4.35 & 4.95 \\
\hline
\textbf{GADS-Hybrid} & \textbf{0.05} & 4.16 & 5.61 & 3.11 & 4.29 & 4.09 & 7.05 & 5.01 & 5.38 \\
\textbf{GADS} & \textbf{0.02} & \textbf{3.61} & 5.05 & 3.04 & 3.90 & 3.84 & 7.06 & 5.00 & 5.30
\end{tabular}

\end{table*}
\begin{table*}[h!]
\centering
\caption{Mean Absolute errors (MAE) and number of parameters for Protocol 2: trained on $70\%$ of the BIWI data-set and tested on $30\%$ of the BIWI data-set (values of SOTA methods are obtained from the study LwPosr~\citep{lwposr}). Methods are sorted in the descending order based on the number of parameters.}
\label{tab:Protocol:2}
\begin{tabular}{c|c|cccc}
\hline
\textbf{Method} & \textbf{Param $10^6$} & \textbf{Yaw} & \textbf{Pitch} & \textbf{Roll} & \textbf{MAE} \\
\hline
DeepHeadPose & - & 5.67 & 5.18 & - & - \\
Martin & - & 3.60 & 2.50 & 2.60 & 2.90 \\
VGG16-RNN & 138.00 & 3.14 & 3.48 & 2.60 & 3.07 \\
VGG16 & 138.00 & 3.91 & 4.03 & 3.03 & 3.66 \\
TokenHPE & 86.42 & 3.01 & \textbf{2.28} & \textbf{2.01} & \textbf{2.49} \\
6DRepNet & 39.32 & 2.69 & 2.92 & 2.36 & 2.66 \\
TriNet & 26.00 & 2.99 & 3.04 & 2.44 & 2.80 \\
FSA-Caps-Fusion & 1.20 & \textbf{2.89} & 4.29 & 3.60 & 3.60 \\
SSR-Net-MD & 0.20 & 4.24 & 4.35 & 4.19 & 4.16 \\
LwPosr & 0.15 & 3.62 & 4.65 & 3.78 & 4.01 \\
\hline
\textbf{GADS-Hybrid} & \textbf{0.05} & 3.20 & 4.02 & 3.16 & 3.46 \\
\textbf{GADS} & \textbf{0.02} & 3.31 & 5.00 & 2.94 & 3.75
\end{tabular}
\end{table*}

\subsubsection{GADS model architecture}
The input for GADS architecture is the normalized 3D landmarks (x,y and z coordinates of 67 landmarks - vector of $67\times3$). During the normalization process, the landmark of the nose is removed since the normalized value of that landmark becomes 0. Within the architecture, as a pre-processing step, the selected landmarks of the five groups have to be extracted (left eye, right eye, left cheek , right cheek, chin). GADS architecture starts with five parallel layers of Deep Sets for each landmark group, as visualized in ~\Cref{fig:gads_arch}. Input for each landmark group is the landmark vector of that group (Ex: for $1^{st}$ deep set layer, input is the landmarks of the left eye in 3D, $6\times3$ which consist of the $x, y, z$ coordinates of the six landmark points). The hyper-parameters of the Deep Set Block~\citep{dshpe} is as follows:
 \begin{itemize}[noitemsep]
    \item Number of encoder layers = 1
    \item Number of decoder layers = 1 
    \item Invariant function = ``max"
\end{itemize}
The outputs of the five parallel units of deep set layers are given as the input for the multi-head attention layer. The number of heads in the multi-head attention layer is \textbf{4}. The outputs of the multi-head attention layers (output-shape = 32) are then concatenated. The concatenated output is passed to the final decoder layer block which consist of two pairs of ``Linear layer + ReLU''. Throughout the architecture, the \textbf{encoder and decoder} layers contain fully connected layers (dense layers) with a ``ReLU" activation function based on the number of units intended to be in the encoder or decoder layer block.

\subsubsection{GADS Hybrid architecture}

As depicted in Figure~\Cref{fig:gads_hybrid_arch}, both the RGB image of the cropped face resized to $64\times64$ and the 67 normalized landmarks are required as input for the hybrid model. The GADS workflow in the hybrid model follows the same procedure as the GADS vanilla workflow. The resized RGB image serves as the input to the CNN architecture, as illustrated in Figure~\Cref{fig:cnn_arch}. Subsequently, the concatenation layer combines the output of both architectures, followed by three decoder layer units.

In both GADS vanilla architecture and the hybrid architecture, a fully connected layer with an output shape of three serves as the final output layer, providing the predicted outputs for the three angles of yaw, pitch, and roll.

\subsubsection{Training specifications}
For both GADS vanilla and hybrid models, we conduct training for 150 epochs with the option for saving the best model. The Adam optimizer is utilized with an initial learning rate of 0.001, and a multi-step learning rate decay is applied at the 60th and 120th epochs, using a decay factor of 0.01. A batch size of 256 with random shuffling is employed for both protocols, and Mean Absolute Error (MAE) is used as the loss function. All hyperparameters of the model architectures were determined based on the results of the ablation study (~\Cref{ablation_study}).

 \subsubsection{Hardware specifications}

PyTorch was used as the implementation framework. A CPU with single-core Intel(R) Xeon(R) @ 2.20GHz and Tesla T4 GPU were used in PyTorch in the Google Colab environment for all experiments in this paper.

\section{Experimental Results}\label{results}
\begin{figure*}[]
    \centering
    \captionsetup{justification=centering}
    \begin{subfigure}[b]{0.3\textwidth}
    \includegraphics[width=\textwidth]{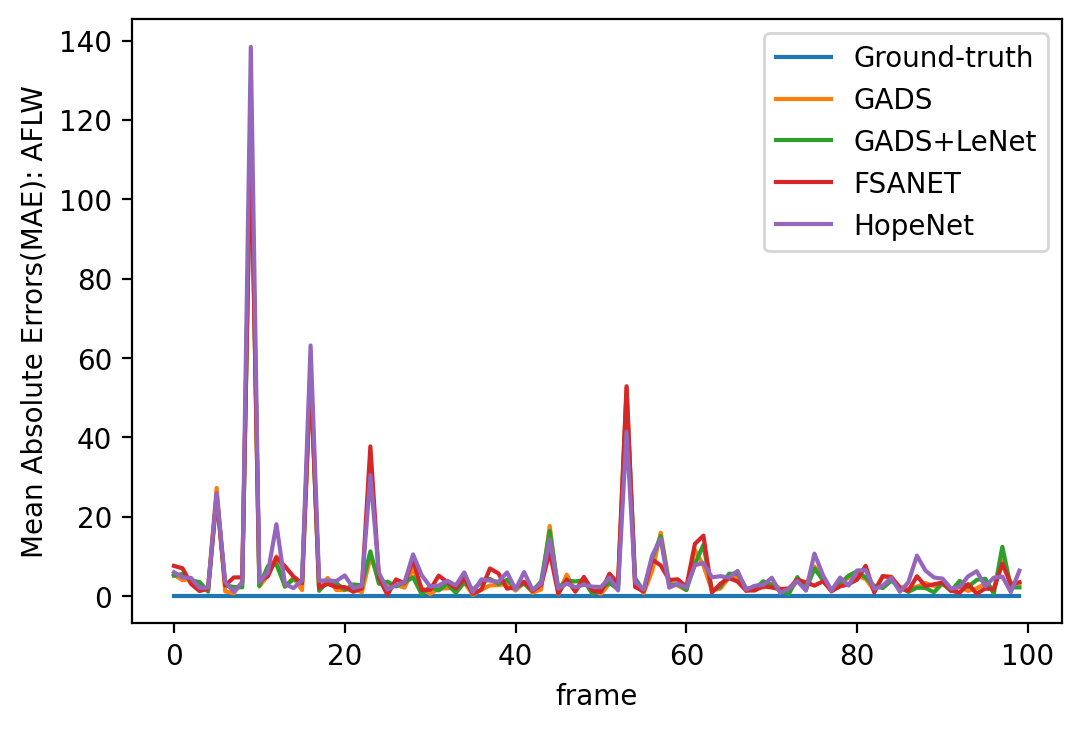}
    \caption{AFLW200 data-set}
    \label{fig:mae_aflw}
     \end{subfigure}
      \begin{subfigure}[b]{0.3\textwidth}
    \includegraphics[width=\textwidth]{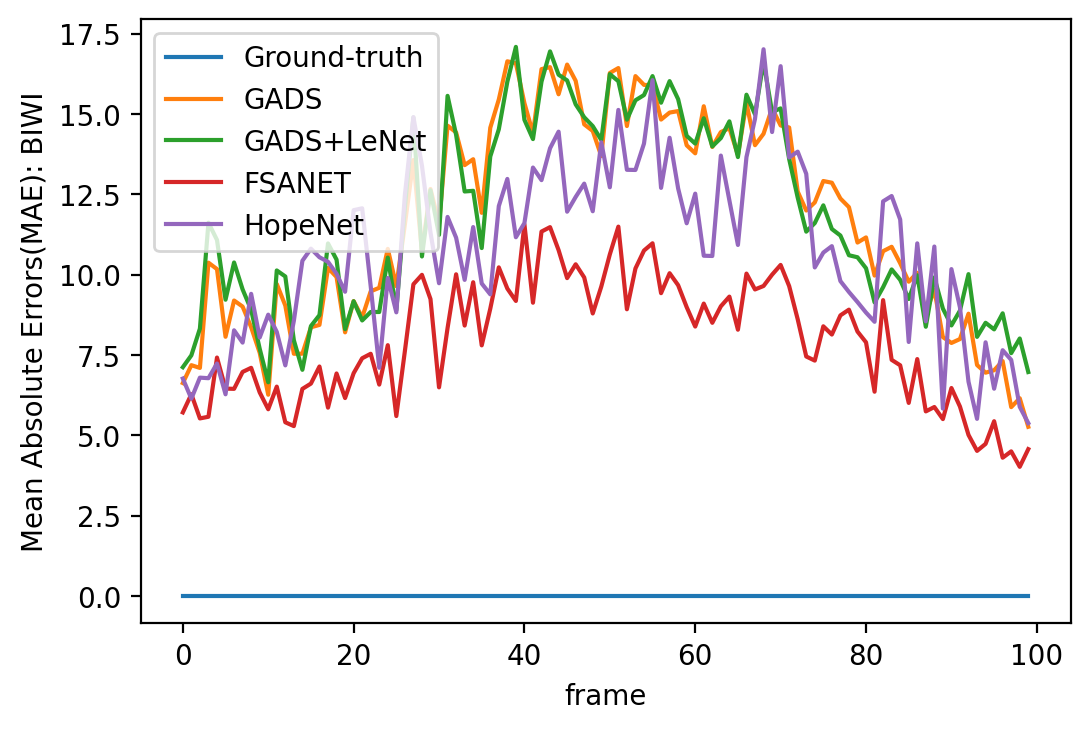}
    \caption{BIWI data-set}
    \label{fig:mae_biwi}
     \end{subfigure}
      \begin{subfigure}[b]{0.3\textwidth}
    \includegraphics[width=\textwidth]{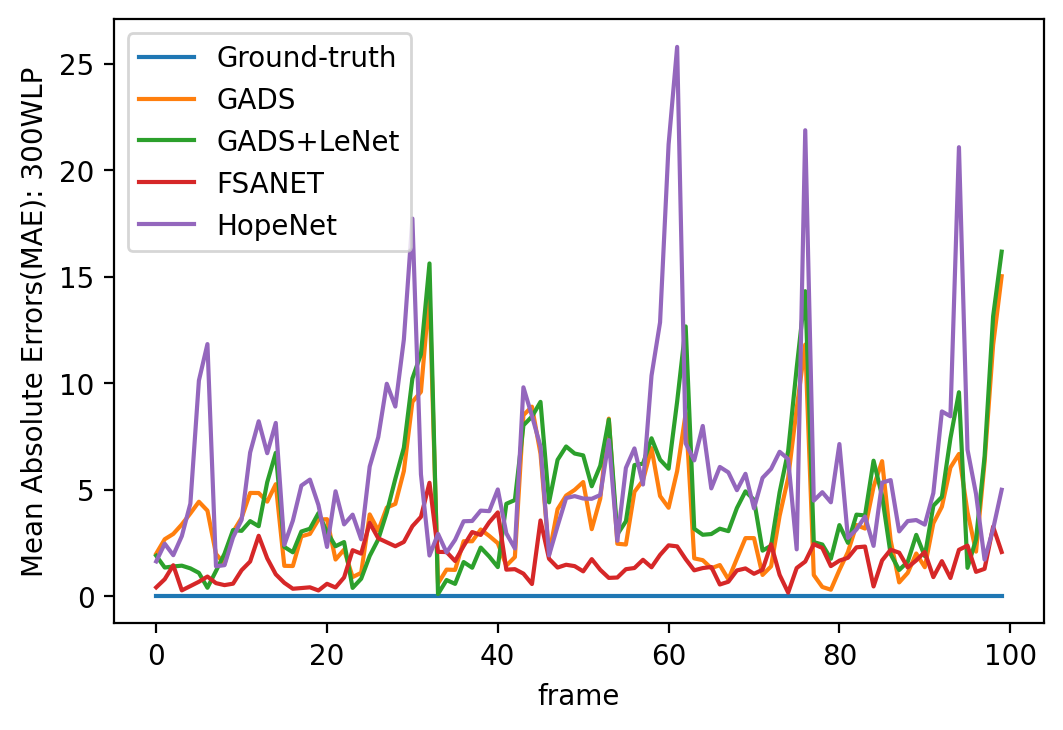}
    \caption{300W-LP data-set}
    \label{fig:mae_wlp}
     \end{subfigure}
 \caption{Mean Absolute error (MAE) of Euler angles variation across 100 samples of the three bench-marking data-sets. }
 \label{fig:maes_all}
\end{figure*}

\begin{figure*}[]
    \centering
    \begin{subfigure}[b]{0.3\textwidth}
    \includegraphics[width=\textwidth]{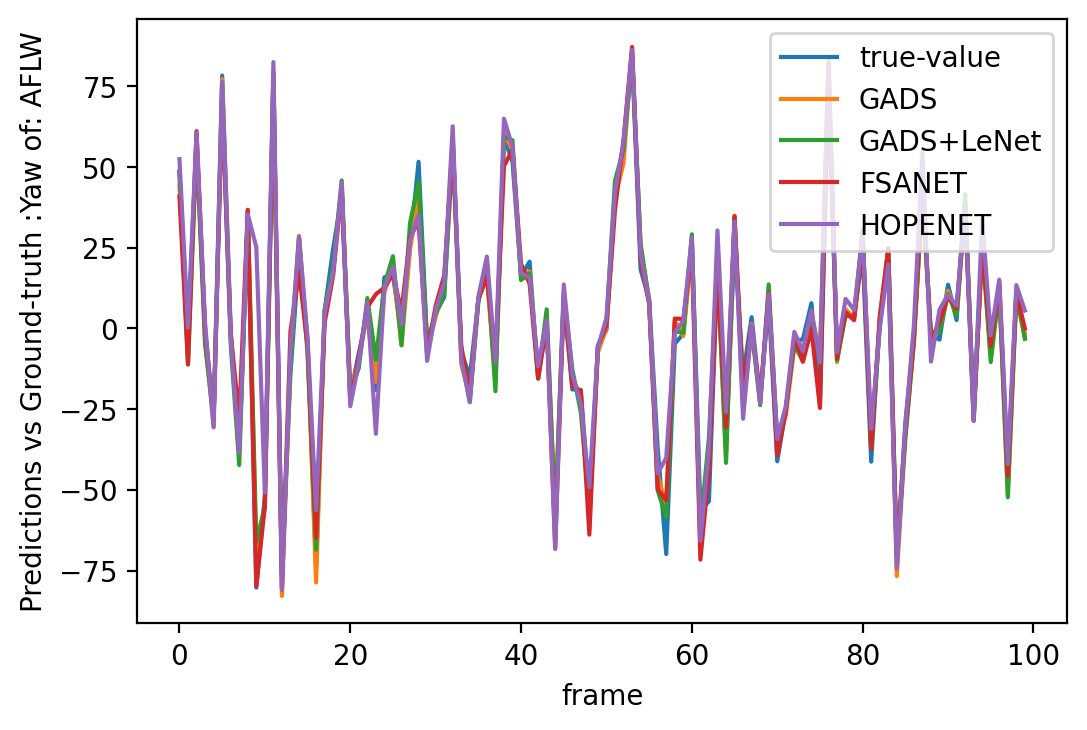}
    \caption{Yaw angle variation}
    \label{fig:yaw_aflw}
     \end{subfigure}
      \begin{subfigure}[b]{0.3\textwidth}
    \includegraphics[width=\textwidth]{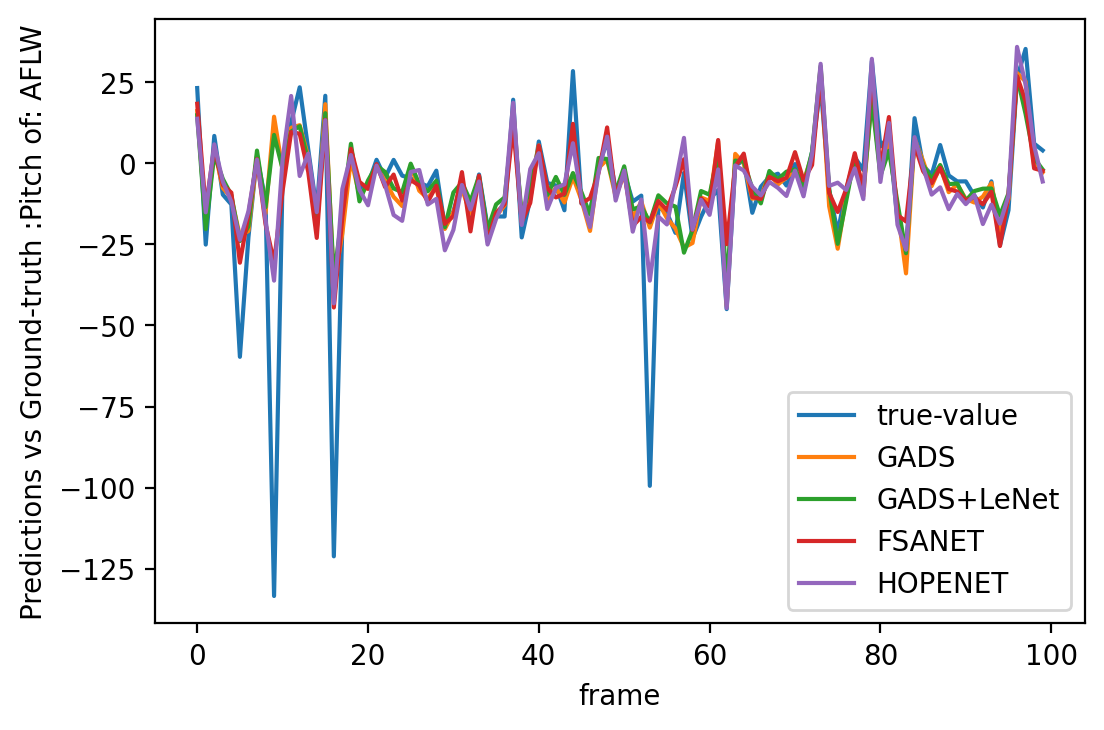}
    \caption{Pitch angle variation}
    \label{fig:pitch_aflw}
     \end{subfigure}
      \begin{subfigure}[b]{0.3\textwidth}
    \includegraphics[width=\textwidth]{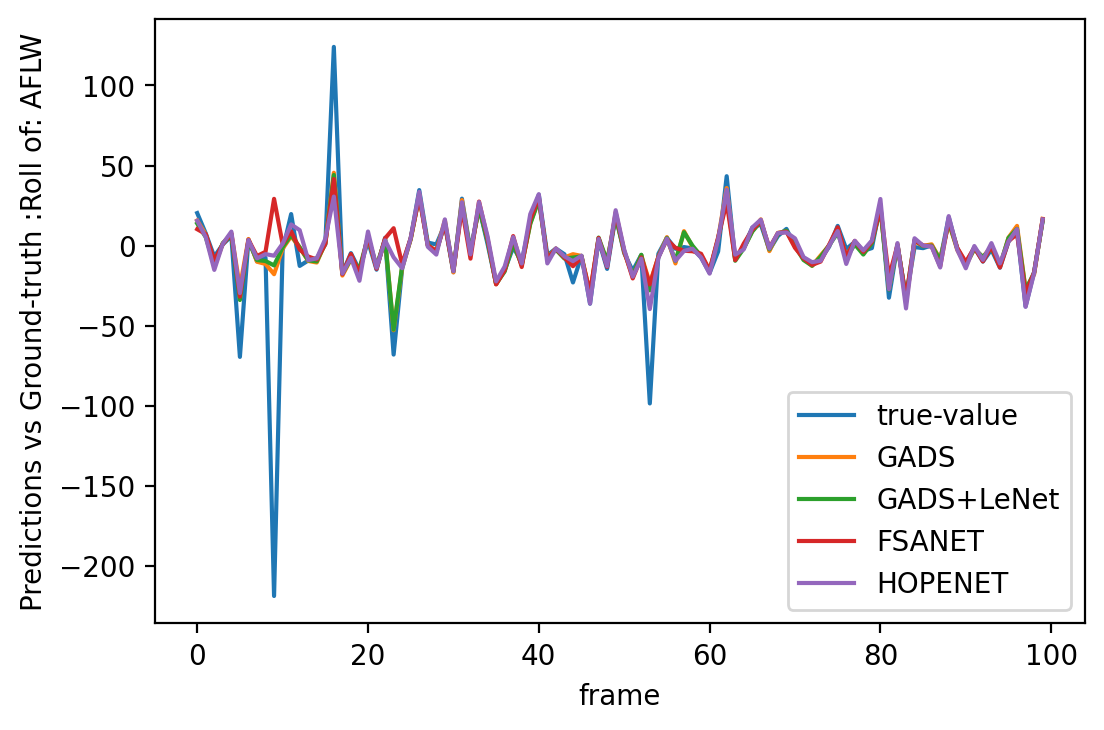}
    \caption{Roll angle variation}
    \label{fig:roll_aflw}
     \end{subfigure}    
 \caption{Variations in Yaw, Pitch, and Roll Euler angles across 100 samples from the AFLW2000 dataset. The graph illustrates the true values, along with predictions from our proposed methods (GADS and GADS-hybrid), as well as predictions from FSANET and HopeNet, each represented in distinct colors. This visualization allows for a closer examination of the alignment between predicted and true values, facilitating a comparison of accuracy among the methods.}
 \label{fig:ypr_aflw}
\end{figure*}

\begin{figure*}[]
    \centering
  
    \begin{subfigure}[b]{0.3\textwidth}
    \includegraphics[width=\textwidth]{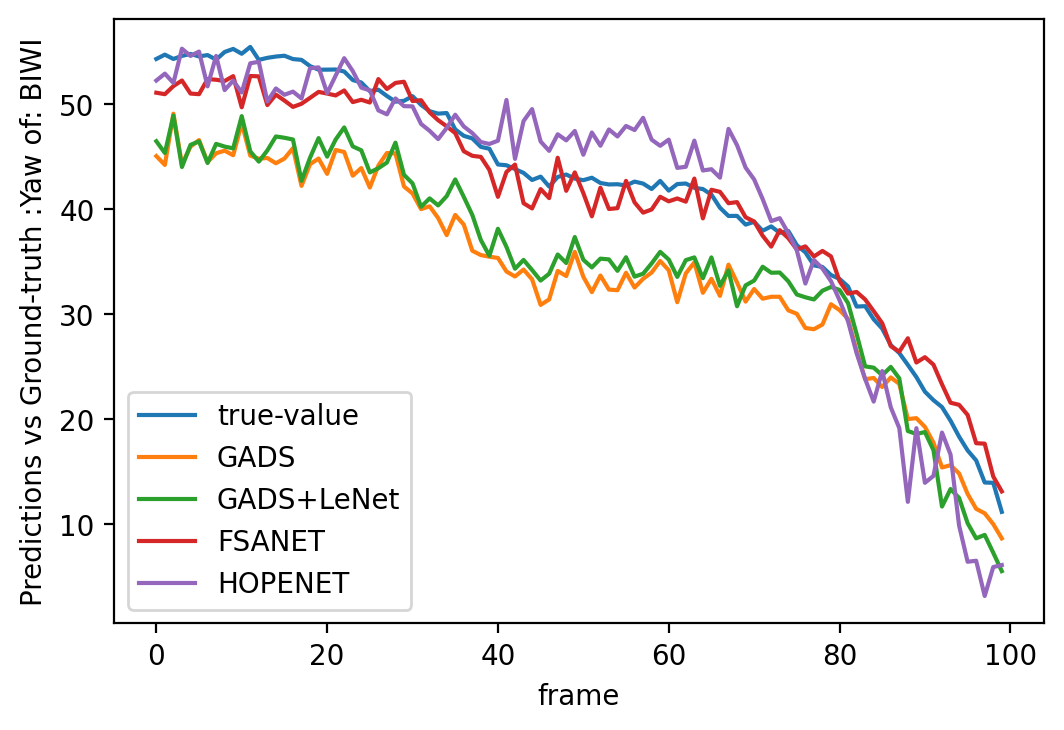}
    \caption{Yaw angle variation}
    \label{fig:yaw_biwi}
     \end{subfigure}
      \begin{subfigure}[b]{0.3\textwidth}
    \includegraphics[width=\textwidth]{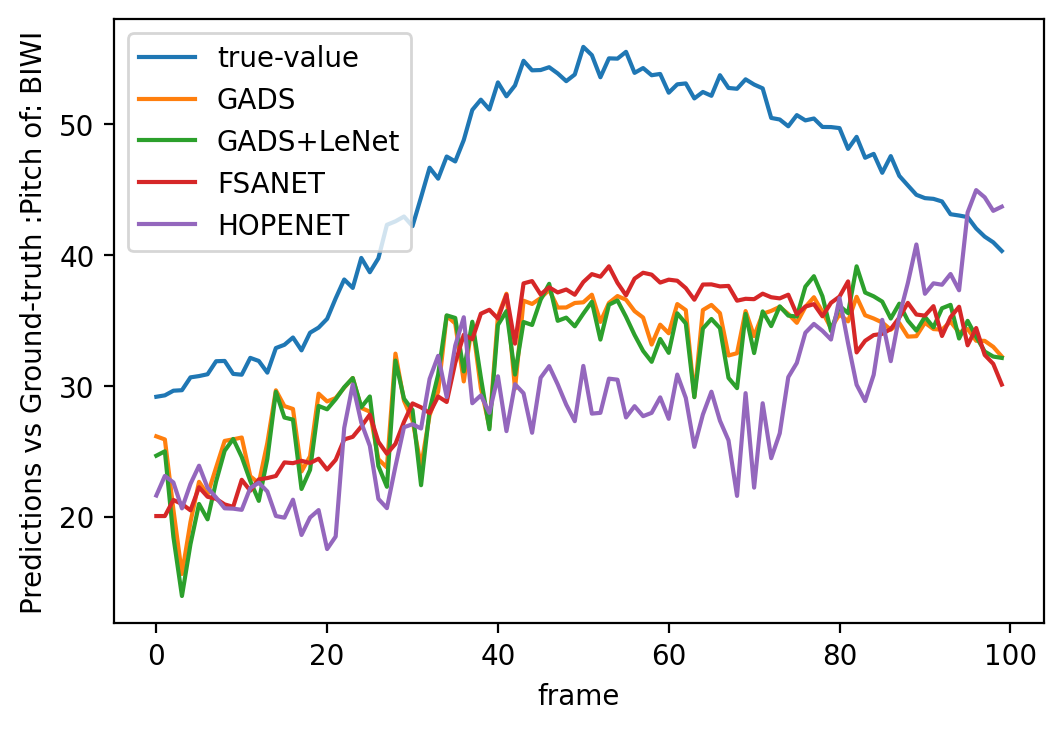}
    \caption{Pitch angle variation}
    \label{fig:pitch_biwi}
     \end{subfigure}
      \begin{subfigure}[b]{0.3\textwidth}
    \includegraphics[width=\textwidth]{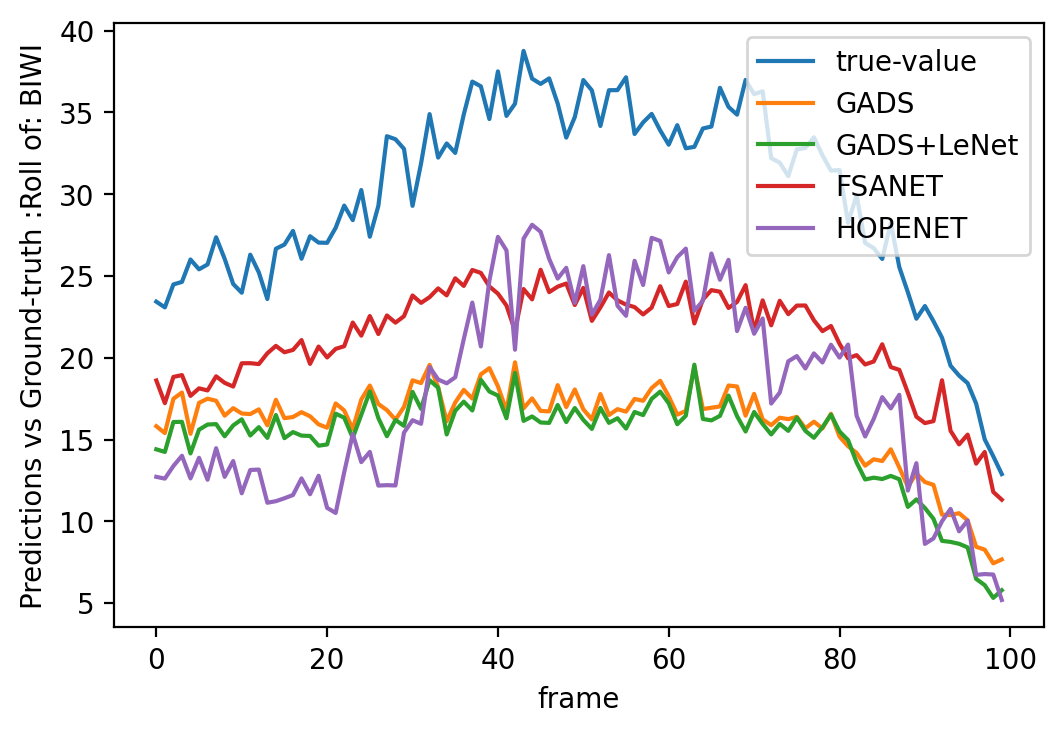}
    \caption{Roll angle variation}
    \label{fig:roll_biwi}
     \end{subfigure}    
 \caption{Variation in Yaw, Pitch, and Roll Euler angles across 100 samples from the BIWI dataset. The graph highlights the differences in angle predictions between our proposed method and two SOTA methods. Notably, all methods show increased discrepancies in predicting higher pitch angles, providing insights into the limitations and performance variations across the models. }
 \label{fig:ypr_biwi}
\end{figure*}
\begin{figure*}[]
    \centering
   
    \begin{subfigure}[b]{0.3\textwidth}
    \includegraphics[width=\textwidth]{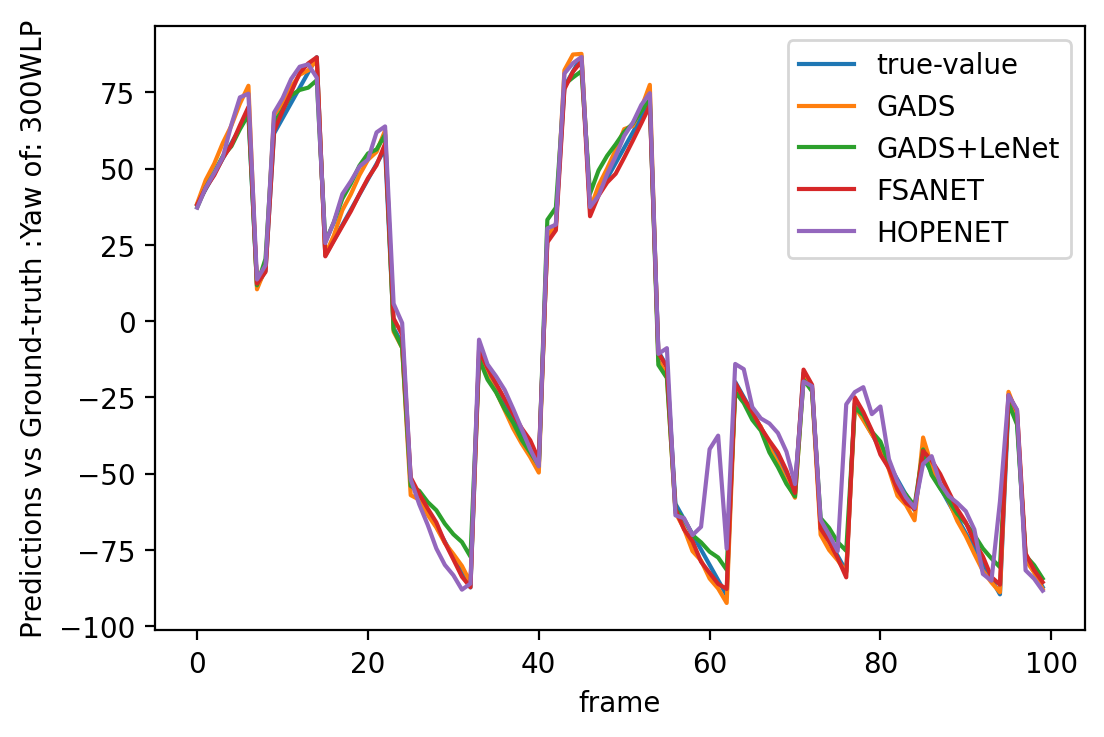}
    \caption{Yaw angle variation}
    \label{fig:yaw_wlp}
     \end{subfigure}
      \begin{subfigure}[b]{0.3\textwidth}
    \includegraphics[width=\textwidth]{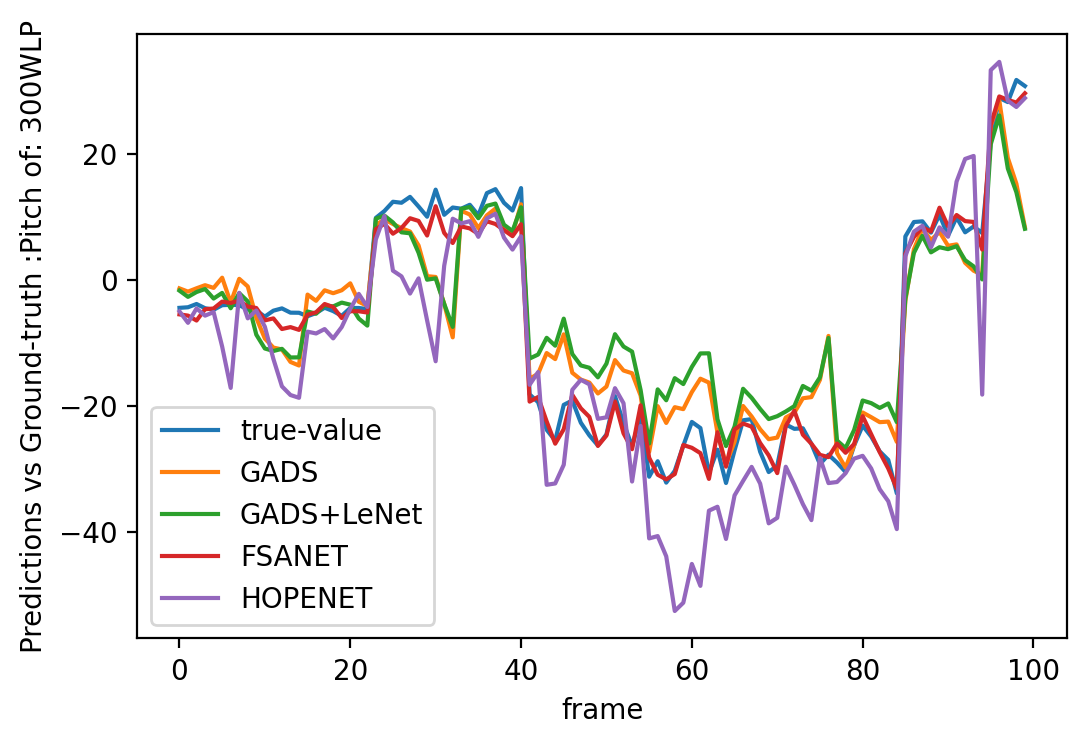}
    \caption{Pitch angle variation}
    \label{fig:pitch_wlp}
     \end{subfigure}
      \begin{subfigure}[b]{0.3\textwidth}
    \includegraphics[width=\textwidth]{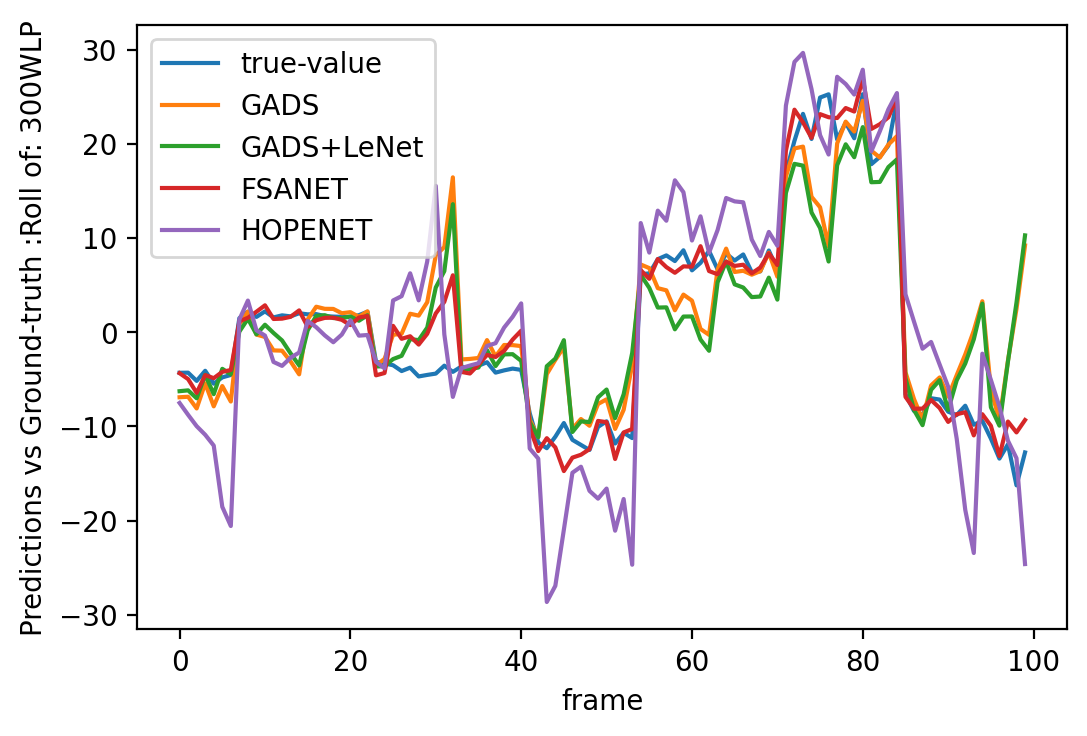}
    \caption{Roll angle variation}
    \label{fig:roll_wlp}
     \end{subfigure}    
 \caption{Variation in Yaw, Pitch, and Roll Euler angles across 100 samples from the 300W-LP dataset. The graph illustrates precise predictions for all three angles, showcasing the accuracy of the models in capturing the variations in the dataset.}
 \label{fig:ypr_wlp}
\end{figure*}
\subsection{Quantitative Results} 
We compare our proposed method for head-pose estimation with state-of-the-art RGB-based methods and Landmark-based methods. Specifically, the SOTA methods are as follows:3DFFA~\citep{3dffa}, KEPLER~\citep{kepler}, FAN~\citep{FAN}, TriNet~\citep{trinet}, Shao~\citep{shao}, HopeNet~\citep{hopenet}, Dlib~\citep{Dlib}, WHENet~\citep{whenet}, EVA-GCN~\citep{evagcn}, FSA-Net~\citep{fsa}, SSR-Net-MD~\citep{ssr}, LwPosr~\citep{lwposr}, DeepHeadPose~\citep{deepHeadPose} Martin~\citep{martin}, VGG16~\citep{vgg16}, 6DRepNet~\citep{rotation} and TokenHPE \citep{tokenhpe}. 

\subsubsection{Results of Protocol 1}
We conducted a thorough evaluation, comparing the number of parameters, absolute errors for yaw, pitch, and roll, as well as the mean absolute errors of the three angles to assess the performance of our proposed method against SOTA methods (~\Cref{tab:Protocol:1}). Notably, our proposed method, GADS, stands out with the fewest number of parameters among existing head-pose estimation methods. The parameter count in GADS is approximately $7.5 \times$ lower than the lightest model, LwPosr~\citep{lwposr}, $220 \times$ smaller than WHENet~\citep{whenet}, $60\times$ smaller than FSA-Net~\citep{fsa}, and notably $4321\times$ smaller than the most recent SOTA method TokenHPE~\citep{tokenhpe}.

Despite it's minimal parameter count, GADS demonstrates competitive accuracy compared to all SOTA methods. Notably, GADS achieves the best mean absolute error in the yaw angle. In the pitch and roll angles, GADS ranks within the top 5, with a marginal decrease in accuracy by 1.11 in pitch compared to TokenHPE~\citep{tokenhpe} and 0.18 in roll. Additionally, GADS exhibits the best accuracy in the yaw angle when tested on the BIWI dataset.

On AFLW2000, GADS exhibits superior performance, particularly excelling in the yaw angle. However, in pitch and roll, GADS ranks within the top 5 in overall accuracy, with a marginal decrease in accuracy by 2.27 and 2.03 for pitch and roll angles, respectively. The average MAE is 1.38 higher than that of EVA-GCN on the AFLW2000 dataset. Notably, our hybrid approach demonstrates slightly lower accuracy than the pure landmark-based method across all segments of Protocol 1. Table \Cref{tab:Protocol:1} includes both Landmark-based and RGB image-based methods. Despite this, our model achieves on-par performance with SOTA methods, showcasing competitive accuracy while maintaining significantly fewer parameters. This characteristic positions our model as highly suitable for edge computing, further emphasizing it's distinction as the fastest among all existing methods.

\subsubsection{Results of Protocol 2}

Based on the results for Protocol 2 (~\Cref{tab:Protocol:2}), our hybrid approach, GADS + LeNet architecture, secures a position within the top 3 among SOTA models. The GADS-Hybrid architecture is 0.31 less accurate than FSA-Net~\citep{fsa} in yaw, 1.74 less accurate than TokenHPE~\citep{tokenhpe} in pitch, 1.15 less accurate than TokenHPE in roll, and 0.97 less accurate in the overall Mean Absolute Error (MAE) than TokenHPE. Despite TokenHPE emerging as the SOTA method in Protocol 2, our hybrid model stands out as the smallest model in the literature for Protocol 2, being $1728\times$ smaller than the TokenHPE architecture. Therefore, our proposed hybrid method maintains a significant size advantage over all existing methods, showcasing it's efficiency despite the need for both landmarks and RGB images.

In summary, both of our proposed methods exhibit outstanding performance on both protocols, achieving a competitive level of accuracy and emerging as the top-performer in specific segments among all SOTA methods. Notably, our proposed methods maintain a size advantage, being at least $3\times$ smaller than the lightest existing method, LowPosr~\citep{lwposr}. This establishes our methods as the lightest models in the literature, making them highly suitable for edge devices. Their simple architecture with minimal parameters further enhances their edge device compatibility.

\subsection{Execution time analysis}
\begin{table}[h!]
\centering
\caption{Execution time comparison of our proposed method vs SOTA methods on CPU and GPU}
\label{tab:exeution_times}
\begin{tabular}{l|l|l}
\hline
\thead{Method} & \thead{Execution time\\CPU(ms)} & \thead{Execution time\\GPU(ms)} \\
\hline
HopeNet & 155.22 & 15.45 \\
FSA-Net & 91.24 & 73.20 \\
\hline
\textbf{GADS-Hybrid} & \textbf{3.63} & \textbf{2.68} \\
\textbf{GADS} & \textbf{2.04} & \textbf{1.89}
\end{tabular}
\end{table}
As outlined, our two proposed architectures feature an impressively low number of parameters. This reduction in parameters directly translates to decreased execution time or latency when predicting the yaw, pitch, and roll angles for a given set of landmarks. To assess execution time, we conducted tests on our two proposed methods and two SOTA methods—HopeNet and FSA-Net, which also have relatively low numbers of parameters compared to other SOTA models—independently on both CPU and GPU (~\Cref{tab:exeution_times}).

Our proposed methods exhibit a speed advantage, being at least $25\times$ faster on the CPU and $27\times$ faster on the GPU compared to FSA-Net. GADS demonstrates a remarkable speed improvement of $75\times$ on the CPU in comparison to HopeNet. These findings underscore that our models, due to their reduced complexity, result in faster execution times. Moreover, the smaller model size, attributed to the lower number of parameters, enhances the edge-device compatibility of our approach.

When comparing the Landmark-based method of GADS with the hybrid model of GADS, the landmark-only model proves to be both faster and smaller, further emphasizing it's edge-device friendliness. To the best of our knowledge, we assert that both our models stand as the most edge device-capable methods, boasting the smallest size and fastest execution time for head-pose estimation.

\subsection{Visualization of Results} 
\subsubsection{Landmark grouping}
\label{landmark_groups}
As described previously (~\Cref{methdolology}), we partition landmarks into five groups and leverage a limited number of landmarks from each group as follows:
\begin{itemize}[noitemsep]
    \item left eye 6 points
    \item right eye 6 points
    \item left cheek 5 points
    \item right cheek 5 points
    \item chin 5 points
\end{itemize}
Out of 68 landmarks we employ total of 27 landmarks divided into 5 groups and the landmark groups are visualized across the AFLW2000 data-set in ~\Cref{fig:land_sets}.

\subsubsection{Ours vs SOTA performance}

\begin{table*}[h!]
\centering
\caption{Ablation Study on GADS Vanilla model architecture}
\label{tab:albation}
\begin{tabular}{c|c|cccc|cccc}
\hline
\textbf{} & \textbf{} & \multicolumn{4}{c}{\textbf{BIWI}} & \multicolumn{4}{|c}{\textbf{AFLW2000}} \\
\hline
\textbf{} & \textbf{Param $10^6$} & \textbf{Yaw} & \textbf{Pitch} & \textbf{Roll} & \textbf{MAE} & \textbf{Yaw} & \textbf{Pitch} & \textbf{Roll} & \textbf{MAE} \\
\hline
\multicolumn{10}{l}{\textbf{No of decoder layers}} \\
\hline
\textbf{1} & \textbf{0.02} &  \textbf{3.61} &  \textbf{5.05} &  \textbf{3.04} &  \textbf{3.90} &  \textbf{3.84} &  \textbf{7.06} &  \textbf{5.00} &  \textbf{5.30} \\
2 & 0.04 & 3.52 & 5.00 & 3.14 & 3.88 & 3.84 & 7.03 & 4.98 & 5.29 \\
3 & 0.06 & 3.78 & 5.20 & 2.99 & 3.99 & 3.94 & 7.07 & 5.07 & 5.36 \\
4 & 0.08 & 3.95 & 5.04 & 3.15 & 4.05 & 4.01 & 7.07 & 5.07 & 5.38 \\
\hline
\multicolumn{10}{l}{\textbf{No of heads in the multi attention}} \\
\hline
 2 &  0.02 &  3.52 & 5.09 & 3.19 &  3.93 &  3.92 &  7.05 &  5.04 &  5.34 \\
\textbf{4} &  \textbf{0.02} &  \textbf{3.61} &  \textbf{5.05} &  \textbf{3.04} &  \textbf{3.90} &  \textbf{3.84} &  \textbf{7.06} &  \textbf{5.00} &  \textbf{5.30} \\
8 & 0.02 & 3.65 & 5.06 & 3.01 & 3.91 & 3.85 & 7.06 & 4.99 & 5.30 \\
\hline
\multicolumn{10}{l}{\textbf{Final no of encoder layers}} \\
\hline
1 &  0.02 &  3.67 &  5.06 &  2.99 &  3.90 &  3.84 &  7.12 &  5.11 &  5.35 \\
\textbf{2} &  \textbf{0.02} & \textbf{3.61} & \textbf{5.05} & \textbf{3.04} & \textbf{3.90} & \textbf{3.84} & \textbf{7.06} & \textbf{5.00} & \textbf{5.30} \\
3 & 0.02 & 3.50 & 5.17 & 3.24 & 3.97 & 3.87 & 7.06 & 4.99 & 5.31 \\
\hline
\multicolumn{10}{l}{\textbf{Activation function}} \\
\hline
Leaky Relu & 0.02 & 3.47 & 5.15 & 3.07 & 3.90 & 3.87 & 7.01 & 4.93 & 5.27 \\
Gelu & 0.02 & 3.68 & 5.09 & 3.07 & 3.94 & 3.90 & 6.97 & 4.95 & 5.27 \\
Sigmoid & 0.02 & 24.15 & 22.71 & 8.07 & 18.31 & 25.44 & 13.24 & 13.88 & 17.52 \\
        \textbf{Relu} &  \textbf{0.02} &  \textbf{3.61} &  \textbf{5.05} &  \textbf{3.04} &  \textbf{3.90} &  \textbf{3.84} &  \textbf{7.06} &  \textbf{5.00} &  \textbf{5.30} \\
\hline
\multicolumn{10}{l}{\textbf{Loss function}} \\
\hline
\textbf{Mean Absolute error} &  \textbf{0.02} &  \textbf{3.61} &  \textbf{5.05} &  \textbf{3.04} &  \textbf{3.90} &  \textbf{3.84} &  \textbf{7.06} &  \textbf{5.00} &  \textbf{5.30} \\
Mean Squared error & 0.02 & 4.82 & 5.30 & 3.34 & 4.49 & 4.39 & 6.96 & 5.09 & 5.48 \\
\hline
\multicolumn{10}{l}{\textbf{Initial learning rate}} \\
\hline
1e-02 & 0.02 & 4.67 & 5.54 & 3.13 & 4.45 & 4.34 & 6.90 & 5.05 & 5.43 \\
\textbf{1e-03} &  \textbf{0.02} &  \textbf{3.61} &  \textbf{5.05} &  \textbf{3.04} &  \textbf{3.90} &  \textbf{3.84} &  \textbf{7.06} &  \textbf{5.00} &  \textbf{5.30} \\
\hline
\multicolumn{10}{l}{\textbf{Dropout}} \\
\hline
\textbf{0} &  \textbf{0.02} &  \textbf{3.61} &  \textbf{5.05} &  \textbf{3.04} &  \textbf{3.90} &  \textbf{3.84} &  \textbf{7.06} &  \textbf{5.00} &  \textbf{5.30} \\
0.1 & 0.02 & 3.70 & 5.23 & 3.50 & 4.15 & 4.16 & 7.21 & 5.51 & 5.63 \\
0.2 & 0.02 & 5.43 & 8.66 & 4.34 & 6.14 & 6.41 & 8.36 & 8.05 & 7.61\\
\hline
\end{tabular}
\end{table*}
With the goal of comparing the performance of our proposed methods, GADS vanilla and the hybrid approach, we illustrate the predicted yaw, pitch, and roll angles of our proposed methods alongside the predictions of HopeNet~\citep{hopenet} and FSA-Net~\citep{fsa}, as well as the ground-truth angles, on a sample from the AFLW2000 dataset (~\Cref{fig:angles_aflw}).

To further evaluate the capabilities of each model, we plot the Mean Absolute Error (MAE) variation for our methods and the two state-of-the-art (SOTA) methods over a sample of 100 frames from each dataset (~\Cref{fig:maes_all}). Additionally, we visualize the yaw, pitch, and roll angles for the same 100 frames in each dataset, plotting the ground-truth and predicted values for all four models to explore the relationship between overall MAE and angle variation of the frames (~\Cref{fig:ypr_aflw}, ~\Cref{fig:ypr_biwi}, ~\Cref{fig:ypr_wlp}). The results affirm that our proposed methods can predict angles at a comparable level with the two RGB-based SOTA methods, despite being at least $3\times$ smaller.

\subsection{Ablation Study} \label{ablation_study}
To assess the performance of our proposed method under varying architecture parameters, we conducted an ablation study using Protocol 1~(~\Cref{tab:albation}). Several parameters were adjusted, including the number of decoder layers in the Deep Set architecture, the number of heads in the multi-head attention layer, the number of encoder layers in the layer before the final output layer, the activation function used in the fully connected layers, the initial learning rate, and the dropout rate of the dropout layers. All other parameters were kept constant during the ablation study.

\textbf{Number of decoder layers:} The number of parameters in the model architecture doubles when the number of layers in the decoder layer increases. However, the accuracy of the resulting model decreases when the layers exceed 2. While the best accuracy was attained with two encoder layers, opting for one encoder layer allows for a reduction in the number of parameters with only a marginal drop of 0.01 in accuracy.
\textbf{Number of heads in the multi-head attention layer:} The accuracy of the model is lowest when the number of heads is 2. Since accuracies are nearly equal for the number of heads 4 and 8, it's best to pick 4 as it will result in a smaller parameter count.

\textbf{Final number of encoder layers:} The accuracy shows slight variations with the number of layers. For the best accuracy, we selected two layers just before the final output layer.

\textbf{Activation function:} We tried four activation functions for the linear layers of the architecture. Except for Sigmoid, the other three activation functions gave almost the same results with fluctuations less than 0.5.

\textbf{Loss function:} Comparing Mean Absolute Error (MAE) and Mean Squared Error (MSE), the best accuracy was obtained when MAE was used as the loss function. Therefore, we employed MAE as the loss function for both of our approaches.

\textbf{Initial learning rate:} Increasing the initial learning rate to higher values resulted in fast convergence with less accuracy. For the best accuracy, we set the initial learning rate to 0.001.

\textbf{Dropout:} Since the dropout rate reduces the model's accuracy, we did not introduce any dropout into the model's architecture.

We employed these best hyperparameters obtained through the ablation study for both the vanilla GADS architecture and the GADS hybrid model architecture.

\subsection{Strengths and limitations}

Both of our proposed methods demonstrate performance on par SOTA methods. Notably, our GADS vanilla model achieves near SOTA performance with a model size that is $4321\times$ smaller than TokenHPE and exhibits processing speed that is $2\times$ faster than FSANET. This reduction in size is attributed to the simplicity of our model architecture. Furthermore, the GADS hybrid model, while not the smallest, surpasses other SOTA methods in terms of size. A key strength lies in the adaptability of our model for edge computing, particularly in scenarios necessitating on-device edge intelligence for head pose estimation.

Upon comparison with existing methods in the literature, it becomes evident that our model distinguishes itself as the lightest and fastest head pose estimation model while maintaining competitive accuracy with SOTA methods. Despite incorporating both landmarks and RGB images in the hybrid model, it's streamlined architecture enables quicker processing than alternative methods. Consequently, both of our proposed methods offer effective deployment in edge computing environments.

However, our reliance on landmarks in real-world scenarios hinges on the performance of the landmark detector. Failures of the landmark detector at higher angles, stemming from occlusions  or warped images, lead to corresponding setbacks in our method. Additionally, the latency introduced by the landmark detector contributes to the overall inference time in real-world scenarios. These limitations should be considered in the practical application of our proposed methods.

\section{Conclusion}
In this paper, we introduce a novel deep learning architecture called Grouped Attention Deep Sets (GADS) for head pose estimation, leveraging 3D facial landmarks. Additionally, we propose the GADS Hybrid architecture, incorporating both facial landmarks and the RGB image of the face. Our architecture stands out as the lightest model for estimating head pose angles to date, achieving competitive accuracy against SOTA models like TokenHPE, EVA-GCN, FSA-Net, WHENet, and TriNet, despite having a remarkably small number of parameters. The results highlight our model's capability to predict poses with low latency, making it well-suited for deployment on edge devices. Furthermore, our approach facilitates the implementation of downstream tasks reliant on head pose angles in resource-constrained environments. The findings suggest the extensibility of our work to other domains such as emotion recognition, hand-sign recognition, and beyond, offering the potential for deploying lightweight models in resource-constrained settings. We are committed to open sourcing the code and artifacts upon the acceptance of the paper.

\section{CRediT Author Statement}
\textbf{Menan Velayuthan}: Conceptualization, methodology, formal analysis, investigation, visualization, Writing - Original Draft, Writing - Review \& Editing. \textbf{Asiri Gawesha}: Methodology, formal analysis, validation, investigation, visualization, Software, Writing - Original Draft, Writing - Review \& Editing. \textbf{Purushoth Velayuthan}: Formal analysis, investigation, visualization, Writing - Original Draft. \textbf{Nuwan Kodagoda}: Resources, Supervision, Project administration, Writing - Review \& Editing. \textbf{Dharshana Kasthurirathna}: Supervision, Writing - Review \& Editing. \textbf{Pradeepa Samarasinghe}: Project administration, Funding acquisition, Writing - Review \& Editing.

\section{Acknowledgments} 
This study was funded by the World Bank through the Accelerating Higher Education Expansion and Development (AHEAD) Operation of the Ministry of Higher Education of Sri Lanka 

\section{Declaration of competing interests}
Pradeepa Samarasinghe reports financial support was provided by Accelerating Higher Education Expansion and Development (AHEAD) Operation of the Ministry of Higher Education of Sri Lanka. If there are other authors, they declare that they have no known competing financial interests or personal relationships that could have appeared to influence the work reported in this paper.

\section{Ethics approval}
This article does not contain any studies involving human participants performed by any of the authors.

\section{Data availability}
The data-sets analyzed during the current study are available in the following resources: 
\begin{itemize}[noitemsep]
\item BIWI dataset~\citep{BIWI} in \url{https://www.kaggle.com/datasets/kmader/biwi-kinect-head-pose-database},
\item 300W-LP~\citep{300WLP} and AFLW2000~\citep{300WLP} in \url{http://www.cbsr.ia.ac.cn/users/xiangyuzhu/projects/3DDFA/main.htm}.
\end{itemize}


\bibliographystyle{elsarticle-harv}
\bibliography{references}

\end{document}